\documentclass[runningheads]{llncs}
\usepackage{graphicx}
\usepackage{tikz,subcaption}

\begin{document}
\title{Towards Hybrid Embedded Feature Selection and Classification Approach with Slim-TSF}
%
%
\author{Anli Ji\inst{1} \and
Chetraj Pandey\inst{2} \and
Berkay Aydin\inst{3}}
%
%
\institute{Georgia State University \\
\email{\{aji1\inst{1},cpandey1\inst{2},baydin2\inst{3}\}@gsu.edu}}
\maketitle              
\begin{abstract}
Traditional solar flare forecasting approaches have mostly relied on physics-based or data-driven models using solar magnetograms, treating flare predictions as a point-in-time classification problem. This approach has limitations, particularly in capturing the evolving nature of solar activity. Recognizing the limitations of traditional flare forecasting approaches, our research aims to uncover hidden relationships and the evolutionary characteristics of solar flares and their source regions. Our previously proposed Sliding Window Multivariate Time Series Forest (Slim-TSF) has shown the feasibility of usage applied on multivariate time series data. A significant aspect of this study is the comparative analysis of our updated Slim-TSF framework against the original model outcomes. Preliminary findings indicate a notable improvement, with an average increase of 5\% in both the True Skill Statistic (TSS) and Heidke Skill Score (HSS). This enhancement not only underscores the effectiveness of our refined methodology but also suggests that our systematic evaluation and feature selection approach can significantly advance the predictive accuracy of solar flare forecasting models.

\end{abstract}
\section{Introduction}
Solar weather events, encompassing phenomena like solar flares, coronal mass ejections (CMEs), solar wind variations, and geomagnetic storms, hold significant importance for Earth’s environment and human technological systems. Among many solar phenomena, solar flares are one of the most intense localized explosions of electromagnetic energy emanating from the Sun's atmosphere. When such energy bursts out, it usually travels near the speed of light ranging from several minutes to hours. It often does not occur alone but alongside other events like coronal mass ejections (CMEs) or solar wind, which can trigger severe geomagnetic storms, extensive radio blackouts on Earth's daylight side, and interfere with delicate instruments onboard near-Earth space equipment. Recent studies have employed physics-based or data-driven models \cite{priest2002magnetic} \cite{shibata2011solar} \cite{Kusano2020} to predict solar flares using data primarily sourced from solar magnetograms \cite{Song2008}. Many of these approaches tend to predict solar flares as a classification problem using point-in-time measurements (where a single time point is applied to represent a single event). Such methods often do not consider the intrinsic temporal evolution nature of data \cite{Georgoulis2012} by evaluating different observations as separate entities, meaning the dynamic essence of flares is usually overlooked.

The characteristics of solar flare evolution are important as they are intricately linked to the dynamic behavior of solar active regions, as delineated in prior research \cite{Benz2008} \cite{9671322} \cite{Pandey2023}. Analyzing these temporal characteristics of flares, it becomes possible to reveal potential implicit relationships and capture unidentified patterns between flares and their originating regions. In our prior study \cite{9377906}, we utilized ensembles of interval-based classification models on multivariate time series data for event prediction. However, this method presented a limitation in understanding which features were more pivotal in decision-making and the rationale behind these decisions. Traditional interval-based classifiers often do not support systematic evaluation through random sub-interval sampling, leading to a process where the identification of relevant features (or intervals from the time series) was arbitrarily generated, thereby missing out on extracting meaningful insights from the model. In our subsequent studies \cite{10431590} and \cite{10386908}, we aimed to identify crucial interval features from multivariate time series data using multi-scale sliding windows with varying interval sizes and step sizes as well as an innovative feature ranking schema for identifying feature importance. This advancement seeks to introduce interpretability into previously opaque models, enhancing our understanding of the decision-making processes underlying model prediction. 


In this study, we expand our previous work focusing more on the systematic evaluation of our Sliding Window Multivariate Time Series Forest (Slim-TSF) model. This involves strategically selecting relevant features to enhance our grasp of the temporal dynamics crucial for solar flare prediction. We've introduced an indexing function to improve the model selection process. This function enables us to identify optimal models using a concise set of parameters and features that have shown promise in prior research. Additionally, we employ a customized internal validation schema to cross-verify our findings, ensuring the robustness and reliability of our results. This approach has led to a noticeable improvement in our model's performance. Specifically, we've achieved an average increase of 5\% in True Skill Statistics (TSS) and Heidke Skill Score (HSS) compared to our original Slim-TSF outcomes. This improvement underscores the value of a systematic feature selection and validation strategy in enhancing the accuracy of solar flare predictions.


The rest of the paper is organized as follows: Section \ref{related_work} provides background information on existing time series classification models pertinent to flare prediction. In Section \ref{methodology}, we provide our problem formulation and introduce our multivariate time series classification model and feature ranking method used for extracting relevant feature intervals from provided time series data. Section \ref{experiments} presents our experimental setup and evaluation framework. Finally, Section \ref{conclusions} provides conclusions from our study and discusses potential avenues for future research.

\section{Related Work} \label{related_work}
From the proliferation of available time series datasets \cite{Silva2018} and a wide spectrum of machine learning-based techniques proposed for time series classification, similarity-based and feature-based algorithms are two notable categories utilized for these predictive tasks. Similarity-based methods predict by measuring the similarity between training and testing instances, using metrics like Euclidean distance or Dynamic Time Warping (NN-DTW) \cite{6497440}, \cite{1163055}, \cite{Bagnall2016},\cite{Lines2014}. In contrast, feature-based algorithms generate predictions by extracting temporal features from entire time series or subsequences within them. For solar flare prediction, both full-disk (e.g., \cite{9671322}\cite{Pandey2022} \cite{pandey2023interpretable}) and active region-based (e.g., \cite{9377906} \cite{9750381} \cite{9378006} \cite{Hong2023}) approaches have shown significant impact by utilizing derived time series features.

Using feature-based algorithms that capture associations between target variables and time series instances through derived features, this distinction is particularly evident in tasks like solar flare prediction or other tasks (such as anomaly detection \cite{Homayouni2020}). For example, \cite{10.5555/766914.766918} extracted basic statistical features like mean and standard deviation from global time series to feed a multi-layer perceptron network, though this method neglected localized informative characteristics. In contrast, \cite{Geurts2001} enhanced model interpretability by considering local attributes through piecewise constant modeling and pattern extraction, though it often resulted in simplistic features during selection. Furthermore, \cite{DBLP
/pkdd/EruhimovMT07} incorporated an extensive range of features such as wavelets and Chebyshev coefficients, but this method faced high computational costs and lacked inherent interpretability in high-dimensional data spaces.

It is a challenging task for many feature-based classification methods when dealing with multivariate time series data because they require additional intricate information across features. Such discriminating features are usually hard to generate in high-dimensional space due to the unknown interrelations among input parameters, adding complexity to model construction. To address this problem, various techniques have been attempted to ensemble univariate models from individual feature spaces instead of considering the global correlations between them. These methods focus on extracting relevant features in univariate aspects and then applying traditional machine learning algorithms for classification. Common features include statistical measures (e.g., mean, variance), spectral features (e.g., Fast Fourier Transform coefficients), and time-domain features (e.g., autocorrelation).

For example, Shapelet-based decision trees \cite{Ye2010} combine shapelets (i.e., discriminative subsequences that capture distinctive patterns in time series data) within an ensemble architecture. This method extracts shapelets from the training data and constructs an ensemble of decision trees (e.g., random forest), where each estimator focuses on a different subset of shapelets, typically measured by Euclidean distance. While effective at capturing local patterns in multivariate time series data, this approach can be computationally expensive and may struggle to identify relevant shapelets in high-dimensional spaces that are both informative and broadly applicable. Additionally, shapelets extracted from one dataset might not generalize well to other datasets with different dimensionalities, characteristics, and patterns.

To mitigate these issues, the Generalized Random Shapelet Forest (gRSF) \cite{Karlsson2016} improves upon the original shapelet-based method by measuring distances between randomly selected time series and others within a threshold distance of the representative shapelet. Similarly, the Time Series Forest (TSF) \cite{Deng2013} incorporates subseries, but instead of relying on distance measurements from learned subsequences, it derives summarized statistical features (such as mean, standard deviation, and gradient) within randomly divided intervals of the univariate time series. This approach treats each time step as a separate component and constructs decision trees for each feature dimension to capture temporal relationships and reduce the high-dimensional feature space. However, this method may not fully capture the interrelationships between different components of the time series, leading to a potential loss of crucial inter-channel relationships and dependencies in multivariate data. The Canonical Interval Forest (CIF) \cite{DBLP
/corr/abs-2008-09172} extends TSF by incorporating additional canonical characteristics of the time series and \emph{catch22} \cite{Lubba2019} features extracted from each interval. This approach aims to capture both individual patterns within each time series component and relationships between different components. However, interpreting an ensemble of decision trees remains challenging, making it less intuitive to understand the combined effects of multiple trees on multivariate time series data compared to single decision trees.

Many of these methods focus solely on understanding how each feature behaves independently, without considering interactions between different features. A relationship within a single time series parameter might be significant for one specific feature but not necessarily for others. The connections between distinct features are often unknown upfront. Understanding feature dependencies in time series data is crucial for improving model interpretability and performance \cite{Saeed2023}. Techniques like Partial Dependence Analysis (PDA) are commonly used in quantifying these dependencies but can be challenging to explore and analyze in multivariate aspects. \cite{Angelini2024} proposes a conceptual framework that refines the computed partial dependences on combinatorial feature subspaces (i.e., all the possible combinations of features on all their domains) but still lacks the capability of capturing intercorrelations that differentiate between features. 

In time series classification problems, selecting the most relevant time intervals is crucial when generating features that effectively distinguish our data, thereby ensuring a robust model. However, identifying these relevant intervals is difficult because they cannot be directly determined and typically require a computationally expensive search across the entire series. Extracting the underlying mutual information from these relevant intervals can enhance our understanding of the predictive process and accelerate the transition from research to practical applications in flare forecasting models. Our objective in this work is to establish a framework that can recognize these characteristics and offer deeper insights into the behavior of classifiers during prediction tasks.


\section{Methodology} \label{methodology}
In this section, we will outline our approach, including 
the extraction of statistical features from time intervals, the introduction of the sliding-window time series forest, and the feature ranking technique we employ.

\begin{figure}[tb!]
\includegraphics[width=\textwidth]{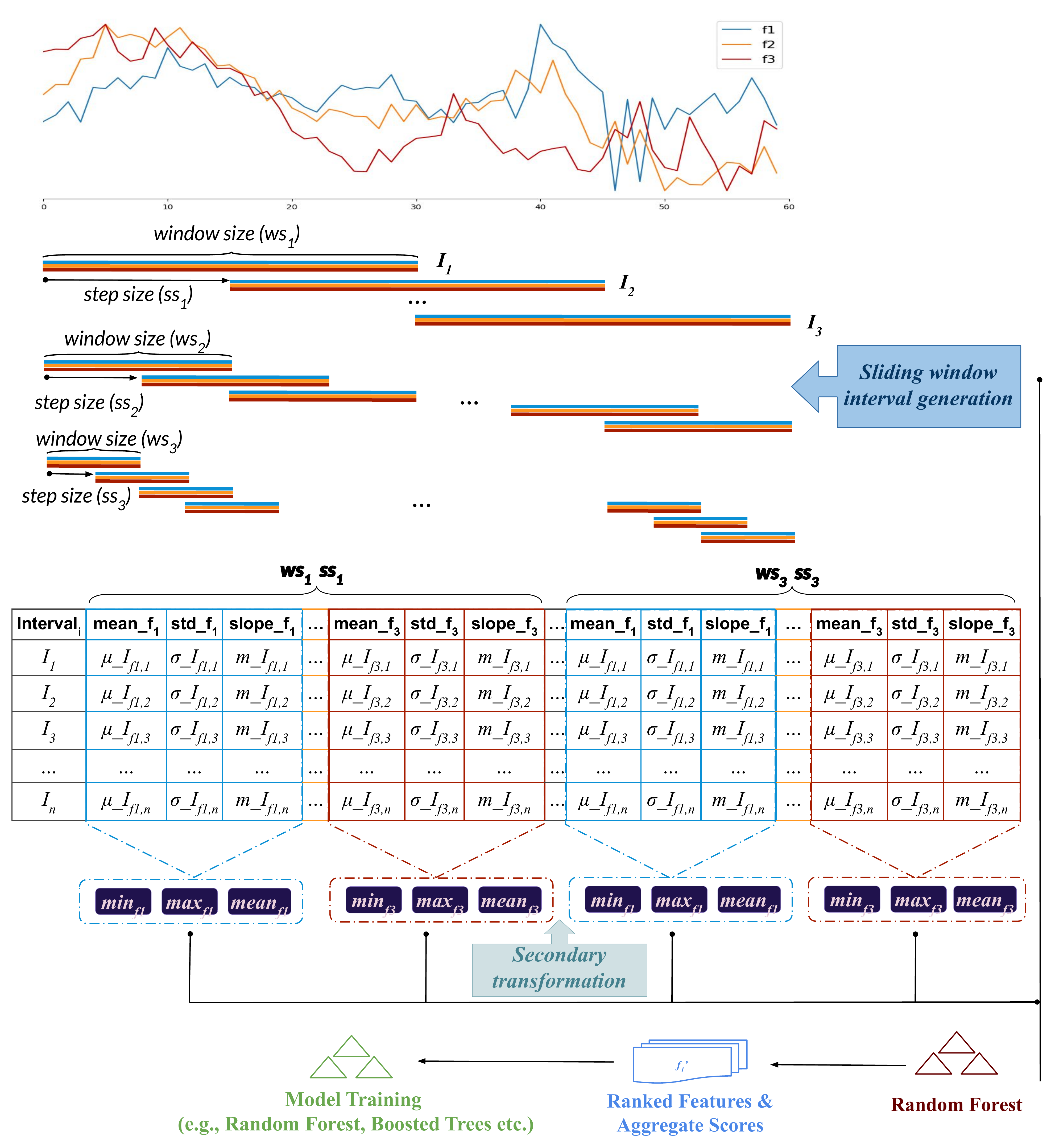}
\caption{An illustration of the sliding window-based statistical feature generation. We first generate subsequences (intervals) with a fixed-size sliding window and step size. Then, we create vectorized features from these intervals where these features can be used as input for the sliding window multivariate time series forest (a random forest built on multivariate time series features) and features are ranked with aggregated relevance scores.} 
\label{fig:overview}
\end{figure}

Our proposed sliding-window multivariate time series forest is an early fusion, interval-based ensemble classification method. Fig.~\ref{fig:overview} illustrates our feature generation process using the sliding window operation. This method employs multiple short decision trees, similar to random forests, which utilize interval-based features extracted from univariate time series through multi-scale sliding windows. By combining features from univariate time series at an early stage, we aim to understand the relationships among these features, using an embedded feature ranking based on mutual information. 



\subsubsection{Interval Features}
To extract well-structured and relevant intervals, we calculate statistical characteristics such as mean, standard deviation, and slope for each interval. Additionally, we derive transformed features like maximum, minimum, and mean through a localized pooling procedure applied to the individual interval features extracted from consecutive intervals after the sliding window operation. In this process, all potential interval sets originating from the same time series are collected, and pooling functions are applied for consolidation. Essentially, we consider the highest, lowest, and average values of statistical properties from each parameter of each subseries obtained through sliding window operations. Formal definitions and explanations for processing multivariate time series and extracting vectorized features and transformation are provided in our previous research \cite{aji2023}.

\subsubsection{\textbf{Sli}ding Window \textbf{M}ultivariate \textbf{T}ime \textbf{S}eries \textbf{F}orest}
After extracting interval features from subsequences obtained through the sliding window operation and applying secondary transformations to these statistical attributes, we merge these two groups of derived features into an input vector. This vector serves as the foundation for creating a versatile time series classifier we refer to as \emph{Slim-TSF}. Among the wide array of supervised learning models available for making predictions, we have chosen random forest classifiers for two reasons: (1) their effectiveness and resilience when dealing with noisy, high-dimensional data, and (2) their ability to select the most relevant features from a given dataset with respect to a target feature.

It is important to highlight that, depending on the chosen parameter settings, such as using smaller window and step sizes, the interval feature vectors' data space can expand considerably. Additionally, the process of vectorization based on the sliding window approach may generate data points that exhibit some degree of correlation and potential noise. Consequently, it is crucial to systematically identify and remove these features. This is achieved through the application of information-theoretic relevance metrics (e.g., Gini index or entropy). This meticulous feature selection process ensures the efficacy of our approach by retaining only the most informative attributes while discarding redundant ones.



\subsubsection{Feature Ranking}
Our ranking methodology involves a systematic process conducted through multiple experiments, denoted by the total number $N$, each executed with distinct experimental configurations. This is analogous to a grid search process. The experiments with different configurations yield individual selected features, denoted as $exp_j$, where $j$ signifies a specific experiment. The ranking denoted as $r$, is a mapping that assigns a rank $i$ to each feature, reflecting its position within the ranking. In each experiment, the features are ranked to create a specific ordering, denoted as $r_{f, i}$, which designates that feature $f$ has achieved the $i^{th}$ rank in that particular experiment. Subsequently, the top-$k$ features selected for inclusion in the selected feature set, denoted as $SFS_j$, within each individual experiment $j$ are determined from the ranking $r$ (i.e., include features whose rank is less than or equal to $k$). This selected feature set is represented as \(\{r_{f, 1}, r_{f, 2},..., r_{f, k}\}\). In the end, the selected feature sets across all experiments are aggregated by summing the sparse representation of top-$k$ membership vectors ($\widehat{SFS_j}$) from each experiment (as in Eq.~\ref{eq4:sfs}).

\begin{equation}\label{eq4:sfs}
SFS = \sum_{j=1,N} \widehat{SFS_j}
\end{equation}

This approach allows for a systematic and consistent method of selecting top features across multiple experiments, enhancing the robustness and reliability of the feature selection process.
Furthermore, we create a counting vector per each interval of each parameter, denoted as $ct_v$ to represent the value counts of individual intervals in the selected feature set $SFS$. This counting vector serves as a transformation function, indicating the frequency with which a given interval appears within the top-$k$ selections of the feature set.

\subsubsection{Hyperparameter Optimization}

Hyperparameter optimization is the process of selecting the optimal hyperparameters to achieve the best performance for a classifier. This process is applied to determine the optimal hyperparameters of our slim-TSF classifier. Traditional grid search cross-validation (CV) is designed for tabulated data and assumes that instances are independent, meaning random assignment of instances to different training and testing folds does not risk overfitting or memorization for the trained models. This can lead to similar instances being included in both training and testing sets, resulting in models that tend to memorize rather than learn. While these models may initially show better results, this is due to sub-optimal sampling rather than stronger generalization capabilities \cite{Ahmadzadeh2019}.

In time series analysis, where instances are obtained with a sliding window, data partitions for training, testing, and validation need to be time-segmented. Traditional grid search CV cannot ensure that instances from consecutive overlapping segments are not placed in both training and testing sets, which undermines the reliability of time series classification performance evaluations. To address this issue, we implemented a customized CV schema that modifies the original grid search to split by the SWAN-SF partitions, maintaining continuous time segmentation instead of randomized sampling. Each time-segmented training partition dataset is assigned a partition label to ensure it is not included in both training and testing sets.

Additionally, we modified the scoring function for our CV, replacing classification accuracy with forecast skill scores, primarily the True Skill Statistic score and the Heidke Skill Score, which will be discussed further in the next section.

\section{Experimental Evaluations}\label{experiments}

The experiments conducted in this study are designed with two primary objectives. Firstly, they aim to demonstrate the effectiveness of time series classifiers developed using distinct interval features and to perform a comprehensive performance comparison among them. Another key objective is to identify the intervals within the time series that hold the greatest relevance to the initial time series. This effort is primarily designed to offer interpretable insights into our model. It involves pinpointing the specific segments of the time series that exert significant influence on predictions and understanding the aggregation strategies that can lead to more accurate outcomes.

\subsection{Data Collection}
For predicting solar flares, we utilized the SWAN-SF dataset, an open-source multivariate time series dataset introduced in \cite{Angryk2020}. This dataset offers a comprehensive collection of space weather-related physical parameters derived from solar magnetograms, incorporating data from various solar active regions and flare observations. Our experiments, encompassing both classification and feature ranking, utilized all 24 active region parameters provided by the dataset. These parameters are widely recognized as highly representative of solar activity, with detailed descriptions available in \cite{Bobra2015} and \cite{Angryk2020}.

The SWAN-SF dataset is organized into five distinct time-segmented partitions to ensure that data instances within each partition do not temporally overlap. Active regions within the dataset are segmented using a sliding observation window of 12-hour intervals across the multivariate time series. Each segment captures essential data, including an active region number (aligned with NOAA Active Region numbers and HMI Active Region Patches identifiers), a class label (indicating the maximum intensity flare expected from that region within the subsequent 24 hours), and start and end timestamps for each segment.

Flare intensity is categorized by the logarithmic classification of peak X-ray flux into major flaring classes (X, M, C, B, or A). For our analysis, instances classified as M- and X-class flares are considered flaring (i.e., positive class), while those classified as C- and B-class flares, along with flare-quiet regions, are treated as non-flaring (i.e., negative class). This binary classification framework is applied to model the solar flare forecasting problem as a binary multivariate time series classification task.

\subsection{Experimental Settings}
To assess our model's performance, we employed a binary $2\times2$ contingency matrix, supplemented by other well-known evaluation metrics for evaluating forecasting accuracy. Within these metrics, the positive class corresponds to significant flare events ($\geq$M1.0 or M- and X-class flares), while the negative class includes smaller flares and regions with minimal flare activity (i.e., instances labeled below the M1.0 threshold). In this context, True Positives (TPs) and True Negatives (TNs) represent instances where the model accurately predicts flaring and non-flaring events, respectively. False Positives (FPs) are false alarms, where the model incorrectly predicts a flare and False Negatives (FNs) are misses, where the model fails to predict an actual flare event.


For rigorous evaluation, we utilize the True Skill Statistic (TSS) and a weighted version of TSS ($\omega$TSS), detailed in Equations Eq.~\ref{eq:tss} and Eq.~\ref{eq:weighted_tss} respectively. The TSS measures the difference between the Probability of Detection (recall for the positive class) and the Probability of False Detection (POFD), providing a robust indicator of model skill.


\begin{equation}
    \label{eq:tss}
        TSS = \frac{TP}{TP + FN} - \frac{FP}{FP + TN}
\end{equation}

In essence, TSS can be reformulated as the sum of true positive rate ($TPR$) and true negative rate ($TNR$), offset by 1 ($TPR+TNR-1$).
The general purpose of TSS is a good all-around forecast evaluation method, especially for evaluating scores among datasets with different imbalance ratios. However, it focuses on a simpler, more understandable scoring schema where both TPR and TNR are treated equally. To change the importance given to each term in this equation we can use a regularization term $\alpha/2$, and create the following weighted TSS ($\omega TSS$):

\begin{equation}
    \label{eq:weighted_tss}
        \omega TSS = \alpha TPR + (2 - \alpha) TNR - 1
\end{equation}

Here, $\alpha/2$ and $1-\alpha/2$ are regularization parameters that show how important each term is. For instance, if correctly predicting an M- or X-class flare is 3 times more important than correctly predicting a non-flaring class, then we can use $\alpha/2=0.75$.

The second skill score we employed is the Heidke Skill Score (HSS), which serves as a critical measure of the forecast's improvement over a climatology-aware random prediction. The HSS ranges from -1 to 1, where a score of 0 indicates that the forecast's accuracy is equivalent to that of a random binary forecast, based solely on the provided class distributions. The formula for calculating this metric is provided in Eq.~\ref{eq:hss2}. Here, $P$ denotes the actual positives, which is the sum of true positives ($TP$) and false negatives ($FN$)), and $N$ represents the actual negatives, the sum of false positives ($FP$) and true negatives ($TN$)).

\begin{equation}
    \label{eq:hss2}
        HSS = \frac{2 \cdot ((TP \cdot TN) - (FN \cdot FP))}{P \cdot (FN + TN) + N \cdot (TP + FP)}
\end{equation}

\begin{figure*}
    \centering
    \begin{subfigure}[b]{0.95\textwidth}
        \centering
        \includegraphics[width=\textwidth]{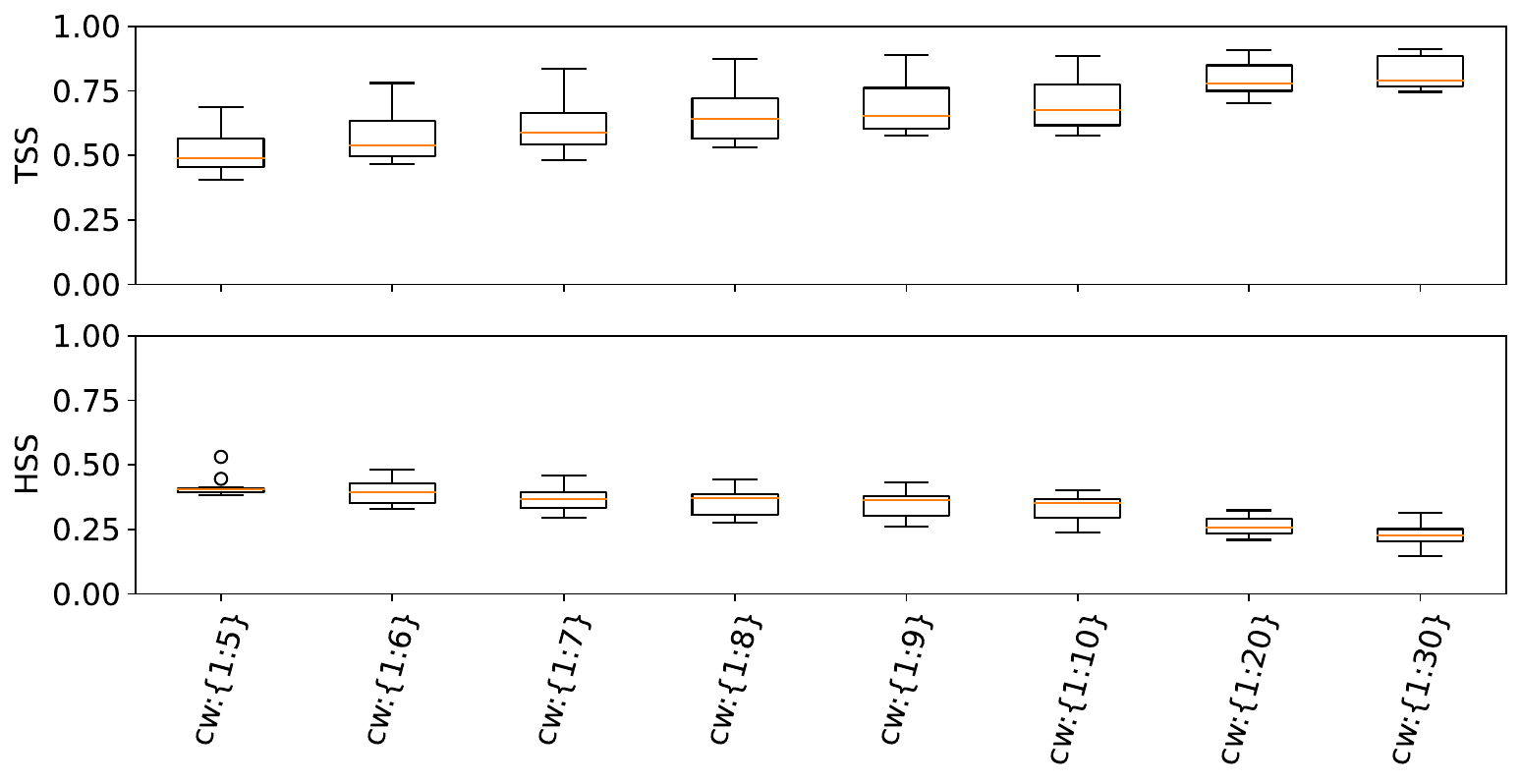}
        \caption{\small Bootstrapping with 10 runs using all derived features}   
        \label{fig:bs10}
    \end{subfigure}
    \hfill
    \begin{subfigure}[b]{0.95\textwidth}  
        \centering 
        \includegraphics[width=\textwidth]{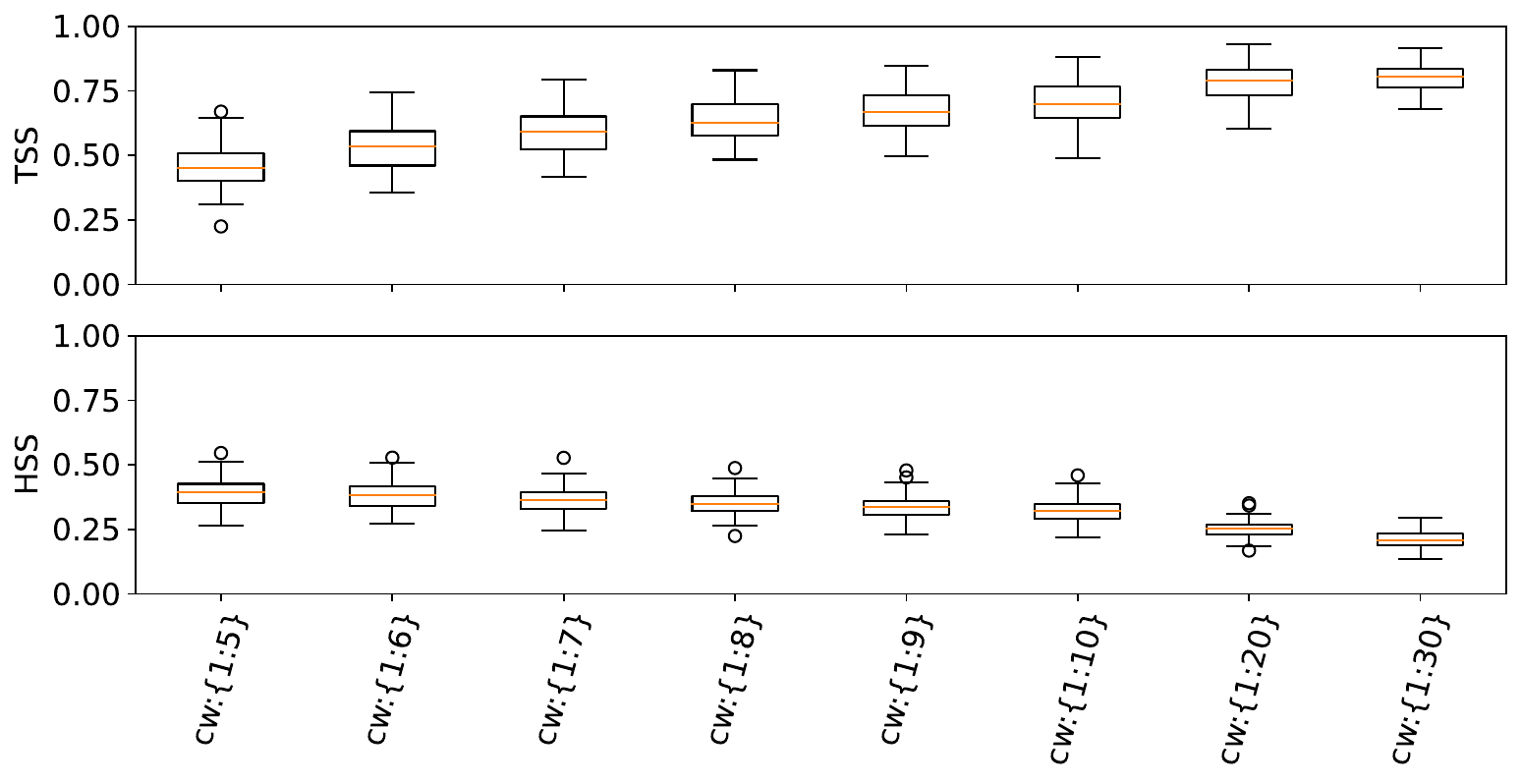}
        \caption{\small Bootstrapping with 100 runs using all derived features}     
        \label{fig:bs100}
    \end{subfigure}
    \hfill
    \begin{subfigure}[b]{0.95\textwidth}  
        \centering 
        \includegraphics[width=\textwidth]{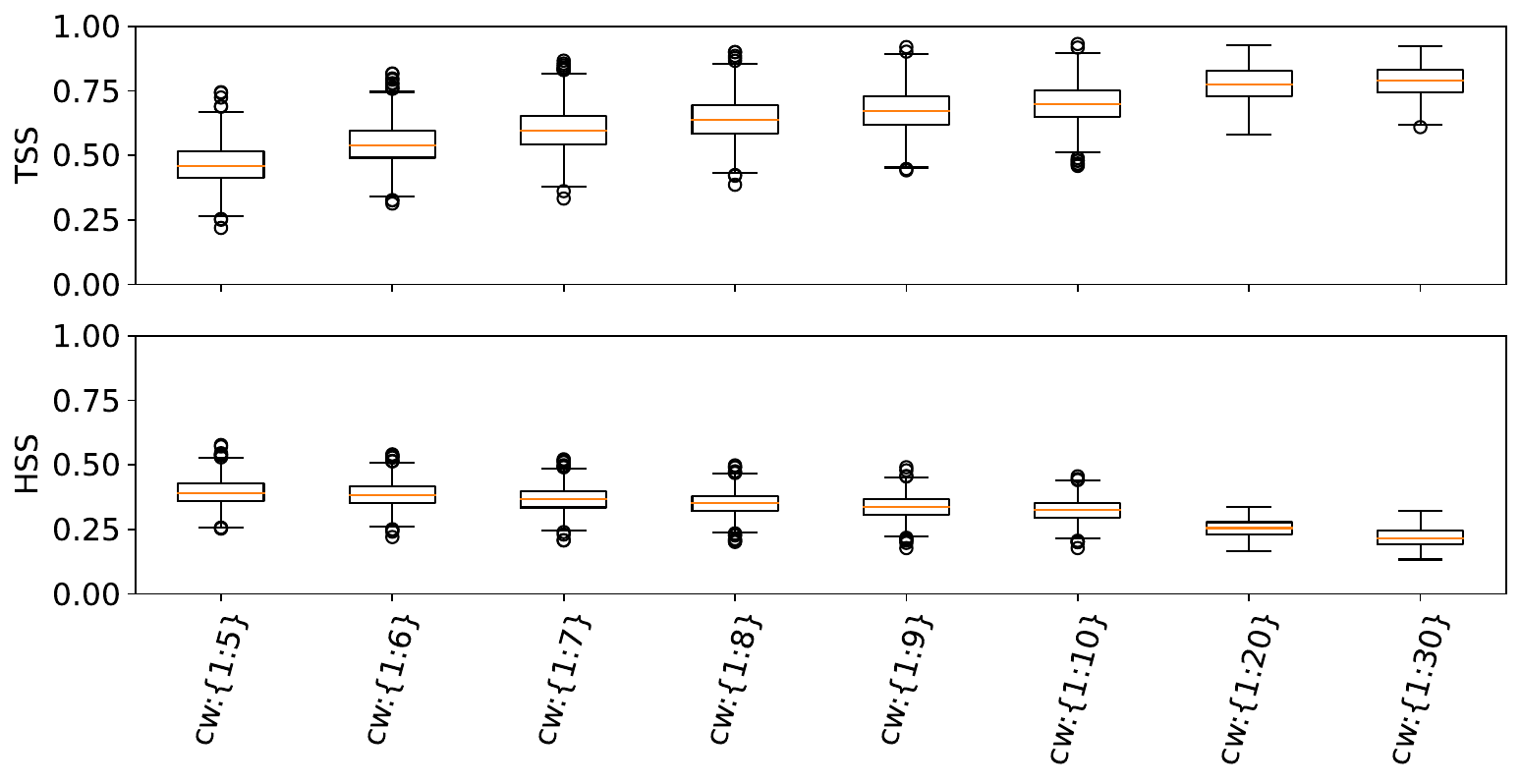}
        \caption{\small Bootstrapping with 1000 runs using all derived features}     
        \label{fig:bs1000}
    \end{subfigure}
    \caption{Error Bar representation of slim-TSF evaluation with ex-ante bootstrapping feature selection using different class weight (i.e., cw) ratio. The most relevant features are selected (per each model trained) across different class weights using the log-scale filter. The TSS and HSS scores are shown for each bootstrapping experiment.} 
    \label{fig:top5_TSS_sub}
\end{figure*}

\begin{figure}[tb!]
\centering
\includegraphics[width=\textwidth]{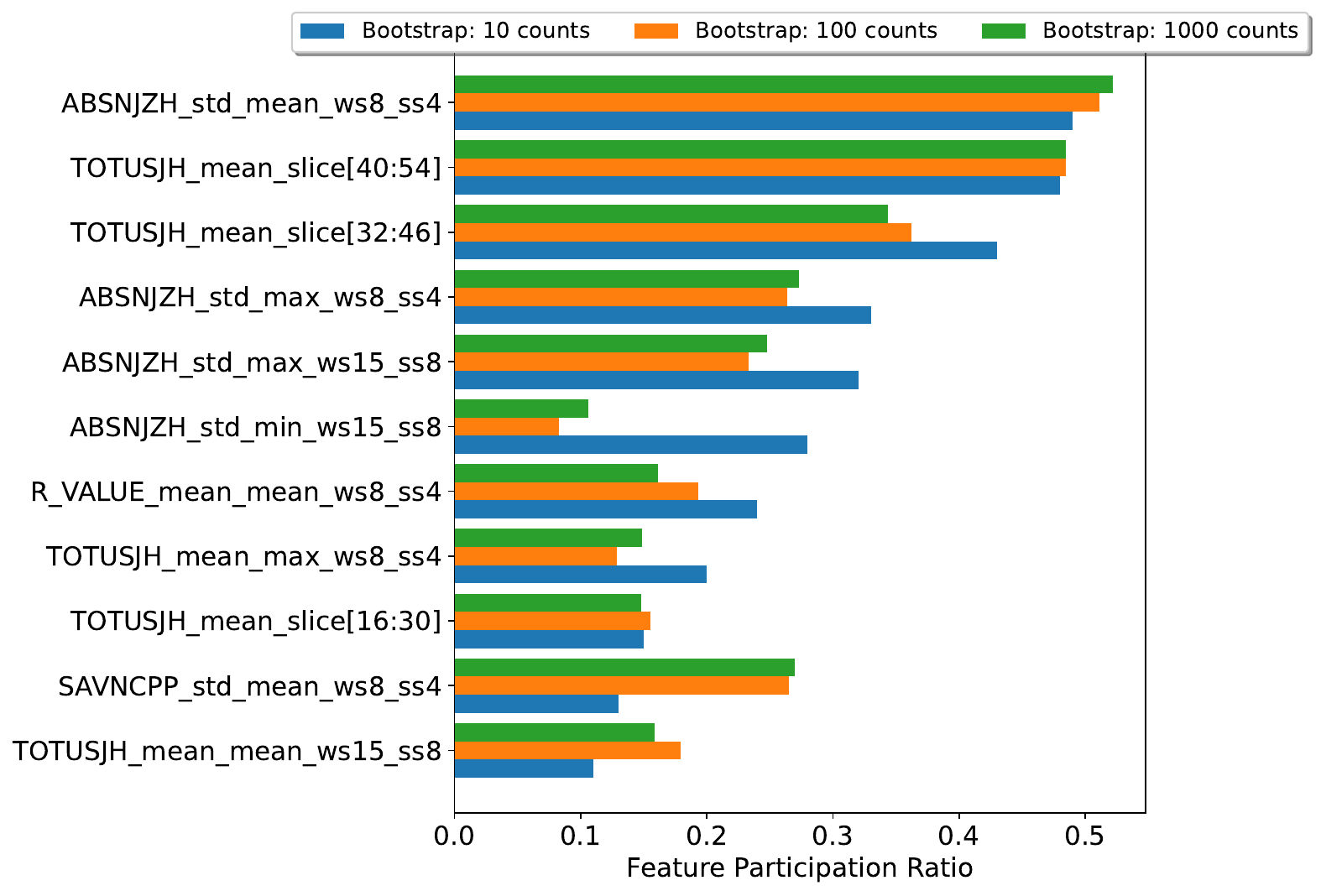}
\caption{A bar plot representation of feature participation ratio in three bootstrap evaluation counts. All features from sliding window intervals and transformed features are used.} 
\label{fig:ratio_plot}
\end{figure}

\subsection{Bootstrapping}
In this study, we introduce a novel approach to feature selection that deviates from the methodology used in \cite{aji2023}. Instead of limiting our selection to only the top 5 highly ranked parameters from individual experiments, we base our feature selection on the cumulative results of the entire bootstrapping process. This involves compiling data from each iteration of the bootstrap subsampling and identifying the most relevant features based on their frequency of appearance throughout the random subsampling procedure. To further refine our feature selection and reduce the risks associated with an overly extensive feature set, we apply a filter $k=log2(N)$ to select the top $k$ features, where $N$ is the total number of features. The results of this refined process are illustrated in Fig. \ref{fig:bs10}, \ref{fig:bs100}, and \ref{fig:bs1000}, demonstrating the use of selected important features from various window and step size configurations. This approach helps to mitigate the influence of outlier features, which might otherwise compromise the accuracy of our predictions.

This refined selection strategy enables us to achieve results comparable to those of our initially proposed Slim-TSF model, but with a significantly smaller set of parameters and features. Consequently, we can maintain average testing scores of approximately 60\% in TSS and 35\% in HSS while utilizing fewer inputs. Despite these reductions, the robustness of our feature selection process is confirmed through its repeated application with random subsamples of the original dataset, ensuring both consistency and reliability. Throughout our studies, certain features, such as those derived from the Total Unsigned Current Helicity (TOTUSJH) and the Absolute Value of the Net Current Helicity (ABSNJZH), are consistently selected across multiple iterations, as shown in Fig.\ref{fig:ratio_plot}. These features have a participation rate of over 40\%, underscoring their critical role in predicting solar flare events.

\subsection{Remarks}
In the results, we demonstrate that the Slim-TSF models, using only the top cumulative important features selected from our bootstrapping iterations, perform comparably, ensuring efficiency and robustness. These models achieve similar outcomes using fewer but more significant features from the original 24 parameters. Notably, models with lower class weights show an average performance improvement of 5\% over previous research \cite{10386908}. This improvement occurs as we reduce redundancy by limiting the use of extensive derived features, thereby increasing feature relevancy. Consequently, the models can concentrate more effectively on key factors by minimizing the redundancy found in less informative features, ultimately enhancing performance significantly.

The outcomes of our study systematically evaluate the performance of our Slim-TSF models, incorporating an additional filter during feature selection. Specifically, our findings reveal that these models improve when utilizing only the top $k$ (from a log-scale) most significant features. This highlights the principle that quality often outweighs quantity in feature selection, as these streamlined models achieve results comparable to their more complex counterparts that utilize all 24 features. Additionally, it is worth noting that adjusting the class weight hyperparameter significantly enhances model performance by reducing the imbalance ratio.

\section{Conclusions} \label{conclusions}
This study builds upon our previous work, which utilized interval-based features generated from sliding window operations in multivariate time series classifiers, also useful for ranking key features, intervals, and transformed pooling features. The primary goal of this work is to enhance the interpretability of high-dimensional multivariate time series classifiers. By employing a comprehensive and methodical approach to feature selection, our research not only improves the predictive accuracy and efficiency of the Slim-TSF model but also offers valuable insights into solar flare prediction, especially under the constraints of limited observational data. This advancement marks significant progress in the field of solar weather forecasting, underscoring the importance of adaptability and innovation in addressing the challenges of data scarcity.

\section*{Acknowledgment}
 This work is supported in part under two grants from NSF (Award \#2104004) and NASA (SWR2O2R Grant \#80NSSC22K0272).

\bibliographystyle{splncs04}
\bibliography{main}
\end{document}